\documentclass[letterpaper, 10 pt, conference]{ieeeconf}  

\IEEEoverridecommandlockouts                              

\usepackage{times} 
\usepackage{amsmath} 
\usepackage{amssymb}  
\usepackage{amsfonts}
\usepackage{caption,subcaption}
\usepackage{hhline}
\usepackage[table]{xcolor} 
\usepackage{tikz}
\usetikzlibrary{arrows, automata}
\usetikzlibrary{shapes.multipart}
\newcommand{\vamp}{\textsc{vamp}}
\newcommand{\dist}{\textrm{dist}}
\usepackage{lipsum}
\usepackage{algorithm}
\usepackage[noend]{algcompatible}
\algnewcommand\RETURN{\State \textbf{return} }
\usepackage{graphicx, caption, subcaption}
\usepackage[sorting=none]{biblatex} 
\usepackage{hyperref}
\hypersetup{
     colorlinks=true,
     linkcolor=blue,
     filecolor=blue,
     citecolor=red,      
     urlcolor=cyan,
     }
\usepackage[compact]{titlesec} 
\titlespacing{\section}{0pt}{2ex}{1ex}
\titlespacing{\subsection}{0pt}{1ex}{0ex}
\titlespacing{\subsubsection}{0pt}{0.5ex}{0ex}

\AtEveryBibitem{
    \ifthenelse{
        \ifentrytype{article}\OR\ifentrytype{inproceedings}\OR\ifentrytype{book}
    }
    {
        \clearfield{url}
        \clearfield{urlyear} 
        \clearfield{review}
        \clearfield{series}
        \clearfield{pages}
        \clearfield{pagetotal}
        \clearfield{issn}
        \clearfield{isbn}
        \clearfield{doi}
        \clearfield{eprint}
        \clearfield{eprinttype}
        \clearfield{location}
        \clearfield{month}
        \clearfield{day}
        \clearfield{series}
        \clearlist{address}
    }
    {}
}

\usepackage{mathtools}
\DeclarePairedDelimiter\abs{\lvert}{\rvert}%

\makeatletter
\let\oldabs\abs
\def\abs{\@ifstar{\oldabs}{\oldabs*}}

\title{\LARGE \bf
Fully Persistent Spatial Data Structures for Efficient Queries in \\Path-Dependent Motion Planning Applications
}

\author{Sathwik Karnik$^{\dagger}$, Tomás Lozano-Pérez$^{\dagger}$, Leslie Pack Kaelbling$^{\dagger}$, Gustavo Nunes Goretkin$^{\dagger}$ 
\thanks{$^{\dagger}$Computer Science and Artificial Intelligence Laboratory, Massachusetts
Institute of Technology,
        {\tt\small \{skarnik, tlp, lpk, goretkin\}@mit.edu}}%
}

\renewcommand{\algorithmiccomment}[1]{\bgroup\hfill\small$\triangleright$~#1\egroup}

\makeatletter
\newcommand{\my@arrow}[1]{\ooalign{$#1-\mkern-5mu-$\cr\hidewidth$#1>$}}
\newcommand{\myarrow}{\mathrel{\mathpalette\my@arrow\relax}}
\makeatother

\begin{document}
\tikzset{bignode/.style={black, draw=orange, fill=yellow!20, minimum size=6em,}}
\tikzstyle{box}=[minimum size = 1.25cm,right=10mm, rectangle, draw=black, fill=white]
\maketitle
\usetikzlibrary{automata,arrows,calc,positioning}
\usetikzlibrary{arrows, automata}
\usetikzlibrary{shapes.multipart}
\tikzset{
  outernode/.style={rectangle split part fill={yellow!20,white},draw,thick, inner sep=0,minimum width = 8cm},
  innernode/.style={inner sep=1em, draw, rectangle split, rectangle split parts=#1}
}
\tikzset{
  outernodepool/.style={rectangle split part fill={green!20,white},draw,thick, inner sep=0},
  innernodepool/.style={inner sep=1em, draw, rectangle split, rectangle split parts=#1}
}

\maketitle
\thispagestyle{empty}
\pagestyle{empty}

\begin{abstract}

Motion planning is a ubiquitous problem that is often a bottleneck in robotic applications. We demonstrate that motion planning problems such as minimum constraint removal, belief-space planning, and visibility-aware motion planning ($\vamp$) benefit from a \emph{path-dependent} formulation, in which the state at a search node is represented implicitly by the path to that node.
A na\"{i}ve approach to computing the feasibility of a successor node in such a path-dependent formulation takes time linear in the path length to the node, in contrast to a (possibly very large) constant time for a more typical search formulation.
For long-horizon plans, performing this linear-time computation, which we call the \emph{lookback}, for each node becomes prohibitive.
To improve upon this, we introduce the use of a fully persistent spatial data structure (FPSDS), which bounds the size of the lookback. We then focus on the application of the FPSDS in $\vamp$, which involves incremental geometric computations that can be accelerated by filtering configurations with bounding volumes using nearest-neighbor data structures.
We demonstrate an asymptotic and practical improvement in the runtime of finding $\vamp$ solutions in several illustrative domains.
To the best of our knowledge, this is the first use of a fully persistent data structure for accelerating motion planning.

\end{abstract}


\section{Introduction}


Motion planning is a crucial computation for many robotic systems that often requires significant resources. Spatial data structures provide asymptotic complexity benefits for many geometric problems, and used judiciously, they also produce practical benefits. In this paper we develop a fully persistent spatial data structure (FPSDS) and explore its use in accelerating motion planning. Before introducing the FPSDS, we discuss two uses of \emph{ephemeral} (not persistent) spatial data structures (SDS) in robotics.

Rapidly-exploring Randomized Tree (RRT) planners typically use a nearest-neighbor (NN) data structure that supports dynamic insertions as planning progresses \cite{yershovaImprovingMotionPlanningAlgorithms2007}. Note that the data structure corresponds to the \emph{algorithmic} state of the RRT search procedure itself, not the state in the sense of the dynamics of the planning problem.

Physics simulations and ray tracing require collision checks between many entities. The entities are approximated with bounding volumes within an SDS (``broad-phase collision checks'') \cite{ericsonRealTimeCollisionDetection2004}.
If the bounding volumes intersect, we say that the entities \emph{interfere}. If two entities do not interfere, then they do not collide.
As the simulation progresses, the bounding volumes are updated.

In both cases, the data structure is dynamically updated \emph{destructively} (i.e. in place) since it is not necessary to access or modify previous versions of the data structure.

There are, however, many important formulations of motion planning problems that benefit from the use of SDS that support access and modification of previous versions.
Like in typical search, e.g. $\mathrm{A}^{*}$ or RRT, these formulations organize search nodes in a tree.
Unlike state-based formulations, these search nodes do not explicitly represent the full state. Computing successor search nodes and their priorities in a queue, in general, requires information not just from a given search node, but the entire path to the root node.
We call these formulations \emph{path-dependent}.
A path-dependent formulation is beneficial in practice for problems where it is more efficient to represent the state implicitly as a sequence, e.g., of previously visited configurations, than it is to represent the state explicitly, e.g., with an occupancy grid or a probability distribution that aggregates information obtained along the path.

Path-dependent formulations lack the property of optimal substructure, without which an optimizing search must, in general, maintain multiple paths to each configuration.
This situation arises when the path cost is non-additive, and efficient solutions are possible in settings with an effective domination criterion~\cite{salzman2017ramp}.
Non-Markovian rewards pose similar challenges~\cite{bacchusRewardingBehaviors1996, gaonReinforcementLearningNonMarkovian2020}.

A straightforward approach to computing node successors in a path-dependent formulation requires performing an operation we call the \emph{lookback} to the root, which requires time linear in the length of the path to the node, whereas in a typical search this operation takes (roughly) constant time.
We can improve upon this by noting that, even though path-dependent formulations generally require information from the entire path, there are applications where only nodes that are local in the workspace are relevant.
Conceptually, we require an SDS at each search node\footnote{The SDS corresponds to the problem state, not the algorithmic state as is the case in RRT}, though in order to benefit (asymptotically and in practice), we must leverage computation and data reuse. Our strategy is to use an FPSDS, which allows access and modification to any version of the spatial data structure \cite{persistence}. For SDSs such as kd-trees \cite{kdtree}, we may use the FPSDS to perform, for instance, range queries against a particular version of the SDS to filter points within a bounding volume. 

In this paper, we describe how to apply this strategy to several problems and describe in detail our contribution of a specific type of FPSDS -- a fully persistent nearest neighbor tree (FPNNT). We describe an application to Visibility-Aware Motion Planning (\vamp{}) \cite{vampwafr}, along with experimental results demonstrating the effectiveness of the FPSDS in illustrative domains.
This application has some overhead, but we show a substantial performance benefit for large domains.


\section{Motion Planning Applications}
\label{sec:example-applications}
Here, we outline three important classes of motion planning problems that benefit from path-dependent formulations.

\subsection{Minimum Constraint Removal}

In the Minimum Constraint Removal (MCR) \cite{hauserMinimumConstraintRemoval2014} problem, the objective is to find a solution that may be infeasible, but that can be made feasible by the removal of a set of constraints. In general, one seeks a solution that requires the removal of as few constraints as possible. Such problems require keeping track of the violations that have been accumulated along each path during the search.

Whereas in shortest-path motion planning, the state is simply the configuration, in MCR, the state of the search problem is the configuration and the set of constraint violations. So, although in shortest-path motion planning, it is sufficient to consider only a shortest path to a given configuration, MCR planning may need to consider multiple paths to a given configuration, each with a different violation set. Whereas the cost function (on transitions) in shortest-path motion planning is additive, in MCR, it is not, because we must not double-count constraint violations.

If there are many removable obstacles, then representing this set explicitly in the state can become prohibitive in memory and in time, since the violation set must be updated non-destructively, and therefore, it must be copied for the new descendants added to the node. For motivation and illustration, consider a variant of MCR, where the objective involves the path swept volume. Given a discretization of the workspace, ``Minimum Swept Volume'' planning is simply MCR, where each workspace cell (voxel) is a removable obstacle. In this setting, it is clear that explicitly representing an occupancy grid at each search node is prohibitive. MCR is amenable to the path-dependent formulation since a path of configurations induces a swept region.


Let $\mathrm{n}_{i}$ represent a search node in the search tree. Let
$\abs{S(\mathrm{n}_1, \ldots, \mathrm{n}_{i})}$ denote the size of the swept region induced by the path of configurations represented in $\mathrm{n}_1$ to $\mathrm{n}_{i}$. We are interested in computing this quantity and improving on the straightforward approach of requiring a lookback to the root (i.e. time linear in the path length).

We provide a sketch of the improvement.
For brevity, let
$A_{i} = S(\mathrm{n}_1, \ldots, \mathrm{n}_{i})$
and
$B_{i+1} = S(\mathrm{n}_{i}, \mathrm{n}_{i+1})$. To update the swept volume from $\mathrm{n}_{i}$ to $\mathrm{n}_{i+1}$, we can recursively compute $|A_{i+1}| = |A_i| + |B_{i+1}\setminus A_i|$.
We must therefore compute the incremental swept volume $B_{i+1}$, but since we would like to avoid storing an explicit representation of $A_{i}$ (storing $|A_{i}|$ is fine), we must visit each node and determine the incremental swept volume at that node, and subtract that set from $B_{i+1}$. This requires only temporary use of a workspace occupancy grid. However, this solution still takes time linear in the path~length.

We improve upon this solution by using the FPSDS to compute fewer swept regions along a path. In this case, we do not need to explicitly compute the entire swept volume $A_i$; rather, we only need to compute swept regions for transitions that interfere with $B_{i+1}$. These transitions can be determined with a range query, based on $B_{i+1}$, to the FPSDS that stores a workspace point corresponding to each transition along the path.

\subsection{Belief-Space Planning}
In the previous example, we relied on representing a swept region explicitly as an occupancy grid or implicitly from a sequence of configurations. In a Partially-Observed Markov Decision Process (POMDP) \cite{pomdp}, a belief state can be represented explicitly as a probability distribution or implicitly as a starting belief and history of actions and observations.

In general, only an approximation of an explicit representation of the belief state is possible. When the state includes the position of objects, and there are difficult-to-handle constraints (e.g., objects are known to not penetrate, or observations tell you some region of space is unoccupied), committing to an explicit posterior and using it in a recursive filtering strategy may ``lock in'' errors in the approximate representation \cite{wongNotSeeingAlso2014}. The alternative is to store all observations, and perform inference on that data as needed.

Consider observations that indicate that the position of an object is uniformly distributed within a disk. The posterior can be computed by set intersection of the observation disks. Whereas the sequence of observations has a straightforward representation, an explicit exact representation of the intersection region does not. However, given the observations, we may sample the exact posterior by, e.g., rejection sampling.

Observations are generated by the environment, but during belief-space planning we might approximate by choosing the maximum-likelihood observations \cite{plattBeliefSpacePlanning2010a}.
The belief over object position could then be used to determine the collision-free probability of an incremental motion.
However, only a subset of observations are likely to be relevant to determining such a quantity. Under certain realistic assumptions, a range query to the FPSDS returns the relevant observations.

\subsection{Visibility-Aware Motion Planning}

\begin{figure}[h]
    \centering
  \includegraphics[width=0.5\linewidth]{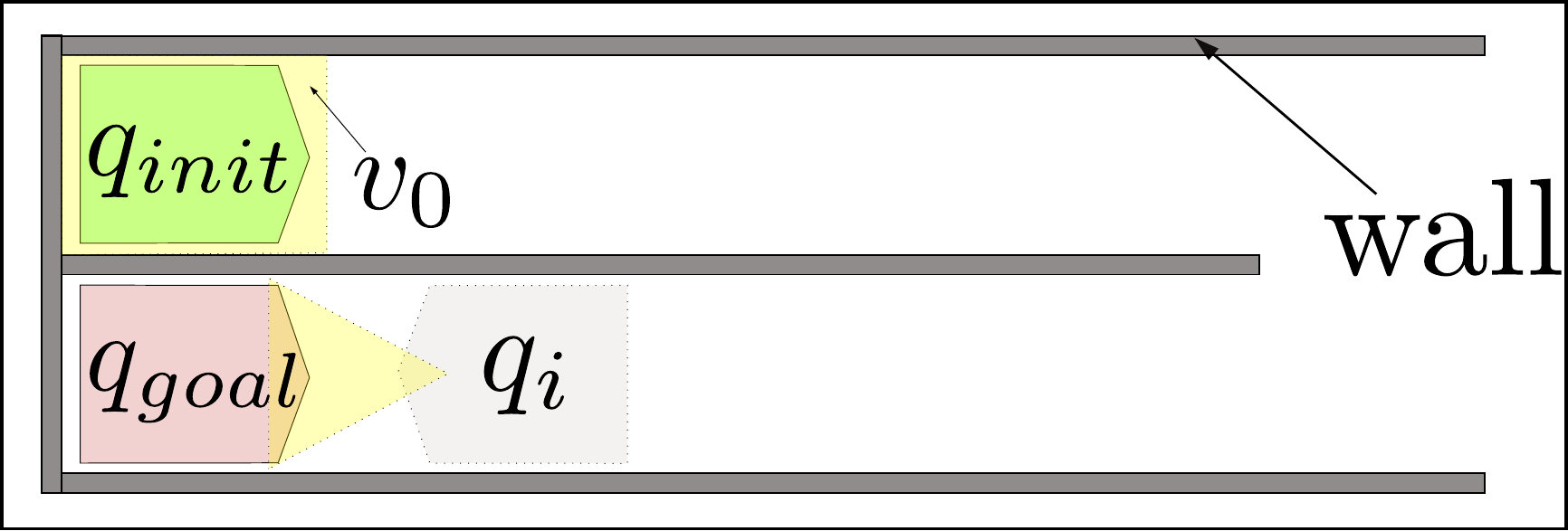}
  \caption{\small Example $\vamp$ problem instance. In the domains we discuss, the viewcone (in yellow) is fixed relative to the robot.}
  \label{fig:horseshoe_diagram}
\end{figure}

In the most classic formulation of robot motion planning, the geometry of the robot and environment is fully determined, and there is perfect actuation, so the problem of finding a path in configuration space that avoids obstacles can be solved open-loop.
We have previously introduced \textit{visibility-aware motion planning}, or $\vamp$ \cite{vampwafr}, where there is uncertainty about the environment in that there may be obstacles not represented in the map given to the motion planner.
The motion planner must produce an open-loop path that avoids all obstacles represented in the map \emph{and} is safe with respect to \emph{unknown} obstacles.
At each moment, the robot may move into a region of space only if there is no obstacle in that region \emph{and} that region was observed (by an on-board sensor) earlier in the path.
The first condition can be determined with collision checking of an articulated body against a static obstacle map. 
The second condition -- called the \textit{visibility constraint} -- requires a different computation and contributes substantially to the planning time. Figure~\ref{fig:horseshoe_diagram} shows an example of a $\vamp$ problem instance in which a violation-free path exists and requires the robot to view the lower hallway before moving backwards into the goal. We focus on algorithmic optimizations to this computation using the fully persistent nearest-neighbor tree (FPNNT).

\section{Fully Persistent Nearest-Neighbor Tree} 
\label{sec:fpsds}
Our contribution of the FPNNT is inspired by the static-to-dynamic logarithmic method \cite{logmethod}, which organizes a forest of static trees (i.e., batch-constructed) and a bounded array (at most size $M$, which is a fixed parameter) to allow dynamic insertions. For $n$ points, there is a static tree that contains $2^{i} M$ points if and only if the  $i^{\text{th}}$ bit of the binary representation of $\lfloor (n-1)/M\rfloor$ is ``1,'' and $(n-1)\pmod{M} + 1$ points that are stored in the bounded array. In the FPNNT, the static trees are nearest-neighbors (NN) \cite{sametDesignAnalysisSpatial1990} trees. The bounded array is encoded across tree nodes, reminiscent of a reverse-linked list, the traversal of which is the lookback.

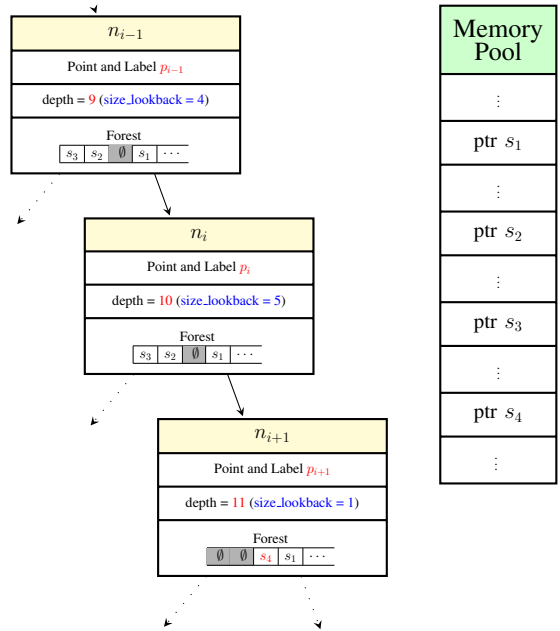
\begin{figure}[h]
\centering
\vspace{0.5em}
\begin{tikzpicture}[
        > = stealth, 
        shorten >=1pt, auto,
    node distance=8cm, scale=0.38, 
    transform shape, align=center, 
    state/.style={circle, draw, minimum size=6ecm}
    ]
    
    \coordinate (a1);
    
    \coordinate (nminus2) at (2.,-6);
    
        \node[outernode]  (nminus1) at ([shift=({290:4 cm})]nminus2)
        [innernode=4] {
            \huge $n_{i-1}$
            \nodepart{second} 
            \Large Point and Label ${\color{red} p_{i-1}}$
            \nodepart{third}  \Large depth = {\color{red} 9} ({\color{blue} size\_lookback = 4})
            \nodepart{fourth} \Large \begin{tabular}{|c|c|c|c|c}
                    \multicolumn{5}{c}{Forest}  \\
                    \hline
                     $s_3$ & $s_2$ & \cellcolor{gray!60} $\emptyset$ & $s_1$ & $\cdots$ \\\hline
                    \end{tabular}
    };
    
    \coordinate[below right of=nminus1] (nleft);
    
    \node[outernode] (n) at ([shift=({290:7.5 cm})]nminus1)
        [innernode=4] {
            \huge $n_i$
            \nodepart{second} 
             \Large Point and Label ${\color{red}p_i}$
            \nodepart{third} \Large depth = {\color{red} 10} ({\color{blue} size\_lookback = 5})
            \nodepart{fourth} \Large \begin{tabular}{|c|c|c|c|c}
                    \multicolumn{5}{c}{Forest}  \\
                    \hline
                     $s_3$ & $s_2$ & \cellcolor{gray!60} $\emptyset$ & $s_1$ & $\cdots$ \\\hline
                    \end{tabular}
    };
    \coordinate (nleft) at ([shift=({230:6 cm})]nminus1);
    
    \node[outernode] (nplus1) at ([shift=({290:7.5 cm})]n)
        [innernode=4] {
            \huge $n_{i+1}$
            \nodepart{second} 
             \Large Point and Label ${\color{red}p_{i+1}}$
            \nodepart{third} \Large depth = {\color{red} 11} ({\color{blue} size\_lookback = 1})
            \nodepart{fourth} \Large \begin{tabular}{|c|c|c|c|c}
                    \multicolumn{5}{c}{Forest}  \\
                    \hline
                     \cellcolor{gray!60} $\emptyset$ & \cellcolor{gray!60} $\emptyset$ & {\color{red}$s_4$} & $s_1$ & $\cdots$ \\\hline
                    \end{tabular}
    };
    \coordinate (nplus1left) at ([shift=({230:6 cm})]n);
    \node[outernodepool,scale=1.3] (pool) at (16.4, -15)
        [innernode=10] {
            \huge Memory \\ \huge Pool
            \nodepart{second} $\vdots$
            \nodepart{third} \LARGE ptr $s_1$
            \nodepart{fourth} $\vdots$
            \nodepart{five} \LARGE ptr $s_2$
            \nodepart{six} $\vdots$
            \nodepart{seven} \LARGE ptr $s_3$
            \nodepart{eight} $\vdots$
            \nodepart{nine} \LARGE ptr $s_4$
            \nodepart{ten} $\vdots$
    };
    \coordinate (nplus2left) at ([shift=({230:6 cm})]nplus1);
    \coordinate (nplus2right) at ([shift=({290:5 cm})]nplus1);
    
    \path[->, loosely dotted] (nminus2) edge (nminus1);
    \path[->, loosely dotted] (nminus1) edge (nleft);
    \path[->, loosely dotted] (n) edge (nplus1left);
    \path[->] (nminus1) edge (n);
    \path[->] (n) edge (nplus1);
    \path[->, loosely dotted] (nplus1) edge (nplus2left);
    \path[->, loosely dotted] (nplus1) edge (nplus2right);
\end{tikzpicture}
\caption{\small Example of part of a fully persistent tree with information stored at each node. The red text at each node denotes the changes from its predecessor node. The subsequence of pointers -- $s_1$, $s_2$, $s_3$, and $s_4$ -- in the persistent tree nodes is shown in the order of allocations in the memory pool. Here, $M=5$.} \label{fig:persist_diagram}

\end{figure}

Each FPNNT node stores (1) a new point and label, (2) a pointer to the predecessor node, (3) the node depth, and (4) a dynamic array of pointers to NN trees (denoted as the \emph{forest}). An example excerpted FPNNT is shown in Figure~\ref{fig:persist_diagram}.

\subsection{Insertion}

Algorithm~\ref{alg:insert_bkd_persistence_node} shows the pseudocode for the function \textsc{Insert\_Node}, which creates a new node of the FPNNT that corresponds with a version of the data structure containing the new point-label pair $p_{succ}$ and all the points in $n_{pred}$.
In the insertion, there are two cases: $d_{pred} \pmod{M} < M$ and $d_{pred} \pmod{M} = 0$. In the first case, the node can simply be inserted. In the second case, we first call the \textsc{LookBack} function, which collects point-label pairs from the $M$ most recent ancestors. In addition, we iterate through the pointers in $f_{succ}$, the forest of NN trees. In particular, we collect the $2^iM$ points stored in each of the trees $T_i$ in the forest for $0\leq i \leq k-1$, where $T_k$ is the first tree that is empty. Altogether these $2^kM$ points are denoted as \text{pt\_labels}. An NN tree is batch-constructed from these points. Finally, we update $f_{succ}$ so that $T_0, T_1,\ldots, T_{k-1}$ are empty and the $T_k$ is the pointer to the newly constructed NN tree. The newly computed $d_{succ}$ and $f_{succ}$ are then used to construct the successor node, which is then inserted into the FPNNT with predecessor $n_{pred}$.


\begin{algorithm}[h]
\begin{algorithmic}[1]
  \State $f_{succ} \gets n_{pred}.\text{forest}$ \COMMENT{Copies predecessor forest}
  \State $d_{pred} \gets n_{pred}.\text{depth}$
  \IF {$d_{pred} \pmod{M} == 0$}
    \State $\text{pt\_labels} \gets  \textsc{LookBack}(n_{pred})$
    \FOR {$i \in 1...\text{length}(n_{pred}.\text{forest})$}
        \IF {$n_{pred}.\text{forest}[i] == \emptyset$}
            \State $\textbf{break}$
        \ENDIF
        \State $\text{pt\_labels} \gets \textsc{Append}(\text{pt\_labels}, f_{succ}[i].\text{pt\_labels})$
        \State $f_{succ}[i] = \emptyset$
    \ENDFOR
    \State $f_{succ}[i] \gets \textsc{Build\_NN\_Tree}(\text{pt\_labels})$
  \ENDIF
  \State  $d_{succ} \gets d_{pred} + 1$
  \STATE $n_{succ} \gets \textsc{Construct\_Node}(d_{succ}, f_{succ}, p_{succ})$
  \end{algorithmic}
  \caption{\textsc{Insert\_Node}($n_{pred}$, $p_{succ}$)}
  \label{alg:insert_bkd_persistence_node}
\end{algorithm}

\setlength{\textfloatsep}{10pt plus 1.0pt minus 2.0pt}

Figure~\ref{fig:persist_diagram} illustrates both cases of the insertion at a given node. The first case is illustrated in the insertion of $p_i$ at node $n_i$ with predecessor $n_{i-1}$. The only changes from $n_{i-1}$ to $n_{i}$ include the new point and label and the size of the lookback. In the second case of insertion, as seen in the insertion of $p_{i+1}$ at node $n_{i+1}$ with predecessor $n_i$, we observe the same changes as in the first case, as well as the new NN-tree pointer $s_4$ in the forest. 

\subsection{Range Query}
 
In the \textsc{Range\_Query} function, we search for the relevant point-label pairs such that the points are within a bounding ball with radius $r_{query}$ of and center $w$. To do so, we query node $n_i$ and then iterate through $f_i$ for a standard range query in each NN tree. Then, we perform a lookback of size $d_i \pmod{M}$ along the ancestors of $n_i$ and perform a brute-force range query against each ancestor.

\subsection{Complexity Analysis}

In this section, we evaluate the time complexities of \textsc{Insert\_Node} and \textsc{Range\_Query}. Assume that the NN trees used in the FPNNT are kd-trees. Let $N$ be the total number of FPNNT nodes, and let $L$ be the length of the longest path in the tree from the root. Let $D$ be the dimension of the points (e.g., the dimension of the workspace). With the motion planning application solutions we seek to optimize, we keep a bounded number of paths to a given configuration. Thus, we can assume that the number of nodes on a given depth in the tree grows polynomially, not exponentially.


\subsubsection{Insertion Time}

For search trees where the number of nodes per level grows polynomially, the time complexity of each insertion is amortized $O((\log L)^2)$.

\subsubsection{Range Query Time}

The range query consists of brute-force querying against the lookback and traditional range querying against each of the kd-trees in the forest. For the lookback, the time complexity is constant, as $M$ is a constant.
For the largest of the kd-trees, the time complexity is worst-case $O(L^{1 - 1/D})$, from the orthogonal range query time complexity \cite{kdtree}. Let $K$ be the total number of points within the radius of the query point. In total, we have a time complexity of $O(L^{1 - 1/D}\log L + K)$, where we use the fact that there are $O(\log L)$ trees. 


\subsection{Comparison with Baseline}

In the baseline method, at each node, we only store a point-label pair and perform brute-force range queries against the path from a node to the root.

\begin{table}[H]
    \centering
    \begin{tabular}{c|c|c}
        Operation                       & Baseline   & FPNNT        \\
       \hline
       \textsc{Insert\_Node}                & $O(1)$                & $O((\log L)^2)$ \\
       \textsc{Range\_Query}        & $O(L)$            & $O(L^{1-1/D}+K)$ 
                
    \end{tabular}
    \caption{\small Amortized time complexities per operation.}
    \label{tab:complexity_comparison}
\end{table}

We can see from Table~\ref{tab:complexity_comparison} that the FPNNT provides a much faster amortized asymptotic runtime in the FPNNT \textsc{Range\_Query} than the baseline \textsc{Range\_Query}. Although this optimization does result in a slower amortized insertion time, in our experiments, we show that this can be a favorable tradeoff. Note that in our planning applications, the number of insertions and queries are roughly equal.

\section{VAMP Problem Formulation}

To observe the benefits of the FPSDS, we focus on the application to $\vamp$. We now provide a formulation of $\vamp$ similar to the one provided in \cite{vampwafr}.

Let $W$ be the workspace ($\mathbb{R}^2$ or $\mathbb{R}^3$), and let $C$ be the configuration space of the robot. Furthermore, let $W_{obs} \subseteq W$ be the region of space known to contain obstacles. Let $q_0$ be the initial configuration and let $v_0 \subseteq W$ be the initial visible region. In this problem, we assume that the entire space swept from the motion of the robot during its path must be previously viewed but the new visible regions are gained only at the end of each primitive motion.

We now define the following functions characterizing visible regions and swept volumes. Let $\mathbb{P}(X)$ denote the powerset of the set $X$. We define $V:C\rightarrow \mathbb{P}(W)$ to be the visibility function -- that is, the function computes the subset of $W$ that is visible from a configuration. We overload the notation and define $V([q_1,\ldots, q_n]) = \bigcup_{i=1}^n V(q_i)$. Let $S:C\rightarrow \mathbb{P}(W)$ denote the swept volume function. As before, we extend this definition so that $S(q_i, q_j) \subseteq W$ represents the space the robot sweeps when moving from $q_i$ to $q_j$. More generally, we define $S([q_1, \ldots, q_n]) = \bigcup_{i = 1}^{n-1} S(q_i, q_{i+1})$. Finally, let $Q_{goal} \subseteq C$ be a set of goal configurations. A $\vamp$ problem instance is represented by the tuple $(W, C, V, S, W_{obs}, q_0, Q_{goal}, v_0)$.

We assume a graph of primitive motions, with vertices embedded in $C$. If there is an edge between $q_i$ and $q_j$, and $S(q_i, q_j) \cap W_{obs} = \emptyset$, then it is collision-free. A path $[q_1, \ldots, q_n]$ is said to be \textit{feasible} if and only if: (1) each edge from $q_i$ to $q_{i+1}$ in the path is collision-free and (2) the path satisfies the \textit{visibility constraint} -- that is, $S(q_i, q_{i+1}) \subseteq v_0 \cup V([q_1, \ldots, q_i])$ for all $i \in \{1, \ldots, n-1\}$.
The state space of this problem is naturally $(q, v) \in C \times \mathbb{P}(W)$, representing the configuration and the visible region attained along the path to the configuration.
As in Section~\ref{sec:example-applications}, the visible region $v_i$ can be represented implicitly by the initial region $v_0$ and the path $[q_1, \ldots, q_i]$. Note that, due to path dependence, the visibility constraint cannot be applied \emph{pointwise} in the same way that the collision constraint can.

\section{Relaxed VAMP Solution}

A feasible solution to \vamp{} can be found in time polynomial in the size of the problem description. We conjecture that the optimal solution is hard, and seek approximate solutions.
The strategy is to relax the visibility constraint and keep track of the \emph{unseen swept region}
$\bigcup_{i=1}^{n} S(q_i, q_{i+1}) \setminus \left(v_0 \cup V([q_1, \ldots, q_i])\right)$, also referred to as the (visibility) violation region. We refer to the computation of the unseen swept region as a \emph{visibility query}.

This relaxed $\vamp{}$ search produces a collision-free path that penalizes, but ultimately allows, visibility violations. As such, it is not responsible for producing a feasible solution on its own, but the violation region is used to inform another search procedure described in earlier work \cite{vampwafr}, which uses the region as a visibility subgoal.
Note that minimizing the size of the violation region does not guarantee that the final path is optimal in length, but it is a useful heuristic for generating effective subgoals.
We will ultimately minimize an over-approximation to the size of the violation region.

\begin{figure}[h]
\centering
\vspace{0.5em}
\includegraphics[width=0.40\linewidth]{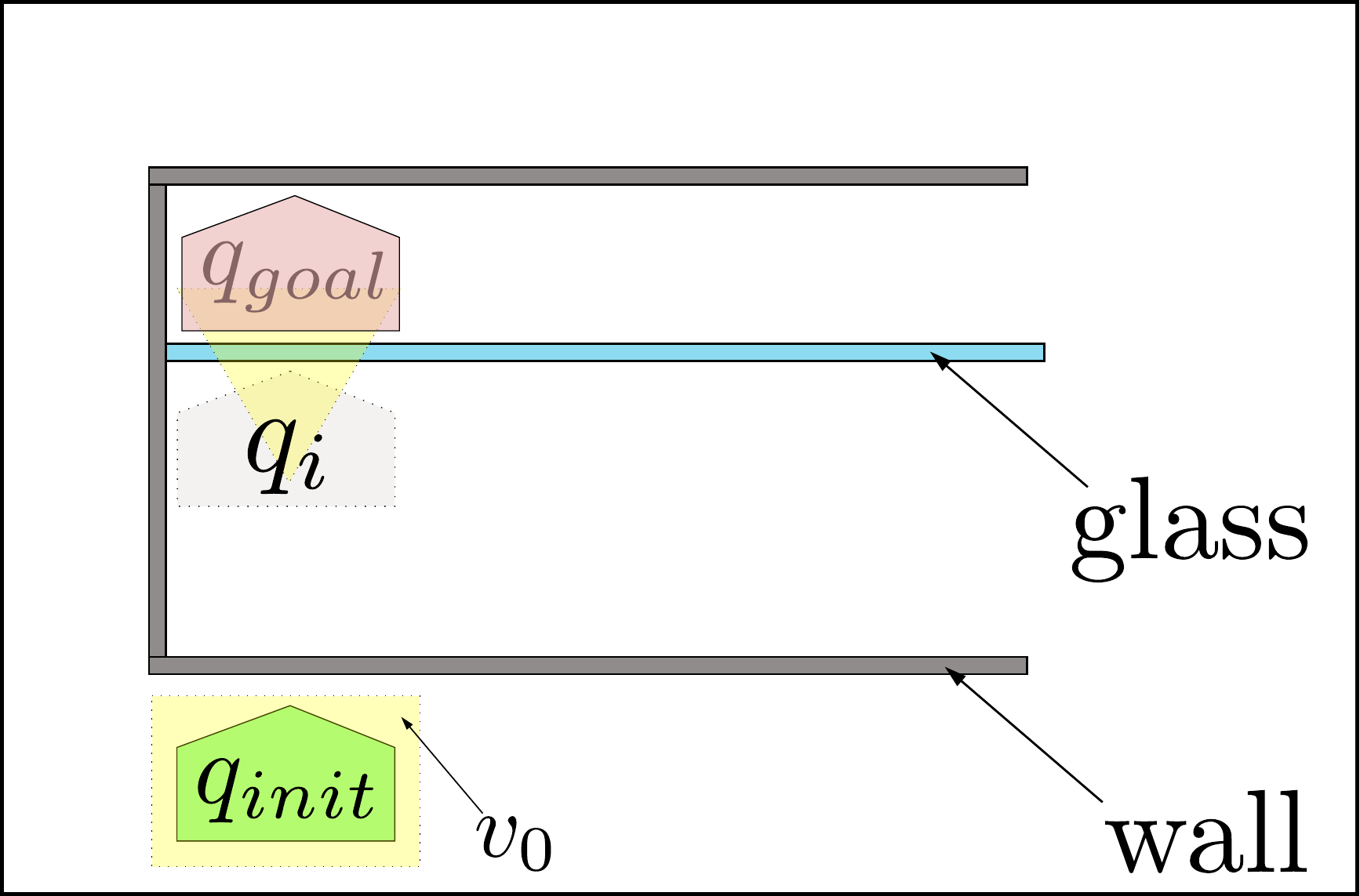}
\caption{\small Motivating example of a $\vamp$ problem instance in which computing the unseen swept region may be expensive. The violation-free path requires the robot to look through the glass wall into the hallway containing $q_{goal}$. The path found from the relaxed $\vamp$ problem results in unseen swept regions.}
\label{fig:glasshallways_diagram}
\end{figure}

We avoid storing a representation of the visible region and violation region at each search node, and determine the incremental violation region, and accumulate its size.
Determining the incremental violation region can be done inefficiently by performing the lookback to the root of the search tree until the incremental unseen region is empty, or until the root node is reached. Due to the relaxation of the visibility constraint, there are some domains where the lookback to the root occurs often (e.g. Figure~\ref{fig:glasshallways_diagram}).

The size of the violation region of an incremental motion -- later described in Algorithm~\ref{alg:find_vis_viol} -- is used to determine the transition cost in the relaxed $\vamp{}$ search (corresponding to the $\textsc{Vamp\_Path\_Vis}$ algorithm with $\mathrm{relaxed = true}$ in earlier work \cite{vampwafr})). This results in an over-approximation of the violation region size, but means that only a subsequence of the search nodes along the path to the root are relevant for determining the new cost, which enables the key optimization of this paper.

\section{Efficient Visibility Queries}

The visibility queries can be made more efficient by filtering out configurations outside of a bounding volume. We formalize the use of bounding volumes in the next section.

\subsection{Bounding Volumes}

Let $\dist(p_1, p_2)$ denote the Euclidean distance between two points $p_1$ and $p_2$. We define a ball $\mathcal{B}(c, r)$ to be the set of points $p$ such that $\dist(p, c) \leq r$. Let $r_{vis}$ be the radius of the smallest ball containing any visible region. We define $\varphi: C \rightarrow W$ such that $\varphi(q) = w_v$, where $V(q) \subseteq \mathcal{B}(w_v, r_{vis})$. Furthermore, define $\psi:C\times C \rightarrow W \times \mathbb{R}_{\geq 0}$ such that $\psi(q_a, q_b) = (w_s, r_s)$, where $r_s$ is the smallest radius such that $S(q_a, q_b) \subseteq \mathcal{B}(w_s, r_s)$. Denote $r_{query} := r_{vis} + r_{s}$.

\begin{figure}[h]
    \centering
    \includegraphics[width=0.5\linewidth]{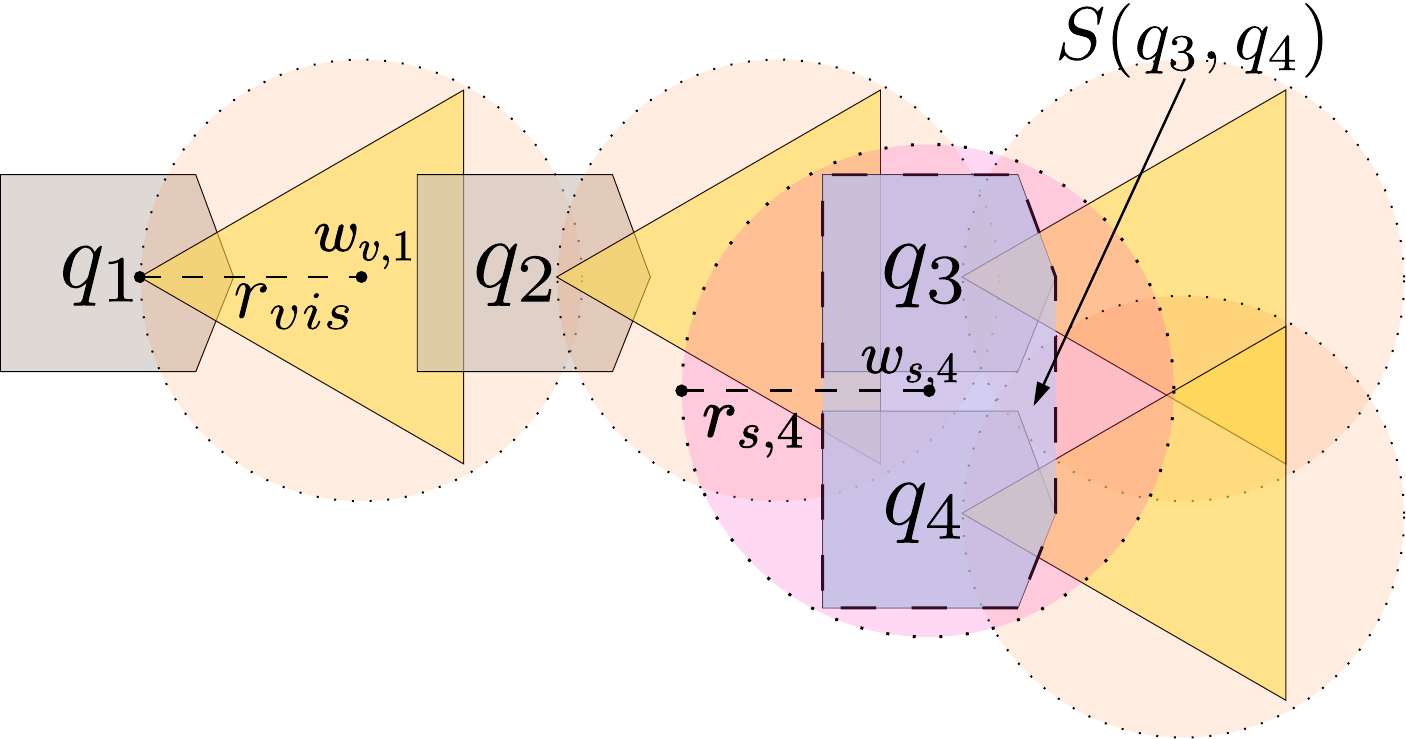}
  \caption{\small Trajectory $[q_1, q_2, q_3, q_4]$ with swept region $S(q_3, q_4)$ from $q_3$ to $q_4$. The viewcones $V(q)$ are shown with bounding radius $r_{vis}$. For $V(q_1)$, we show that the bounding ball has center $w_{v,1}$. The swept region $S(q_3, q_4)$ is shown with a bounding ball $\mathcal{B}(w_{s,4}, r_{s,4})$.}
  \label{fig:swept_region_radius_diagram}
\end{figure}

Consider the path $[q_1, \ldots, q_i]$ in our search tree, and suppose we are interested in calculating the unseen swept volume $S(q_i, q_{i+1}) \setminus V([q_1, \ldots, q_i])$ for some configuration $q_{i+1}$. Let $\psi(q_i, q_{i+1}) = (w_{s, i+1}, r_{s, i+1})$. For $r_{query} = r_{vis} + r_{s, i+1}$, we can guarantee that all configurations $q$ for which $\dist(\varphi(q), w_{s, i+1}) > r_{query}$ have the property that $S(q_i, q_{i+1}) \cap V(q) = \emptyset$. Figure~\ref{fig:swept_region_radius_diagram} illustrates a trajectory with swept region $S(q_3, q_4)$ from $q_3$ to $q_4$. It can be seen that if the distance between the centers of the viewcone bounding balls and $\mathcal{B}(w_{s,4}, r_{s,4})$ exceeds $r_{vis}+r_{s,4}$, the balls do not intersect and, thus, $S(q_3, q_4) \cap V(q) = \emptyset$. 

Since computing the incremental updates to the unseen swept region (see Algorithm~\ref{alg:find_vis_viol} Line~\ref{alg-line:vis-setminus}) can be expensive, our approach considers only the configurations in the path that do not interfere with $\mathcal{B}(w_{s,i+1},r_{s,i+1})$. Algorithm~\ref{alg:find_vis_viol} illustrates this point through the call to $\textsc{Range\_Query}$ in $\textsc{Find\_Vis\_Viol}$, which returns $S(q_i, q_{i+1}) \setminus V([q_1, \ldots, q_i])$. 

\begin{algorithm}
\begin{algorithmic}[1]
  \State $q_i \gets \text{path}[\textbf{end}]$
  \State $(w_{s, i+1}, r_{s, i+1}) \gets \psi(q_i, q_{i+1})$
  \State $r_{query} \gets r_{vis} + r_{s, i+1}$
  \State $\text{unseen\_swept} \gets S(q_i, q_{i+1})$
  \FOR{$q_j \in \textsc{Range\_Query}(\text{path}, w_{s, i+1}, r_{query})$}
    \State $\text{unseen\_swept} \gets \text{unseen\_swept}\setminus V(q_j) $
    \label{alg-line:vis-setminus}
    \IF {unseen\_swept = $\emptyset$}
        \State$\textbf{break}$
    \ENDIF
  \ENDFOR
  \RETURN $\text{unseen\_swept}$
  \caption{\textsc{Find\_Vis\_Viol}(path, $q_{i+1}$, $r_{vis}$)}
  \label{alg:find_vis_viol}
\end{algorithmic}
\end{algorithm}

$\textsc{Range\_Query}$ can be determined by brute force, requiring time linear in the path length. We can improve on this, at the cost of storage, by performing range queries using the FPNNT in the $\vamp$ application. Specifically, we use kd-trees in the  $\textsc{Build\_NN\_Tree}$ function of Algorithm~\ref{alg:insert_bkd_persistence_node}. Using the notation of Section~\ref{sec:fpsds}, we can set the ``points'' as viewcone bounding ball centers $w_{v,i}$ and ``labels'' as configurations $q_i$. Thus, $\textsc{Range\_Query}$ returns the configurations with viewcone bounding balls interfering with the swept region of interest.

\section{Experiments}
We now discuss the experiments that demonstrate that the FPNNT, using kd-trees, provides an efficient way to perform the \textsc{Range\_Query} in Algorithm~\ref{alg:find_vis_viol}. We measure the performance of solving relaxed VAMP problems and compare the FPNNT with the baseline of performing brute-force range queries.
The experiments are implemented in the Julia language \cite{bezansonJuliaFreshApproach2017}. Here, we omit the compilation times.

\begin{figure}[h]
\centering
\begin{tabular}{ccc}
  \includegraphics[height=2.3cm]{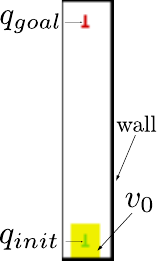} \hspace{0.25cm} &   \includegraphics[height=2.3cm]{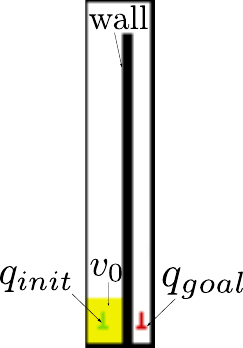} \hspace{0.25cm} & \includegraphics[height=2.3cm]{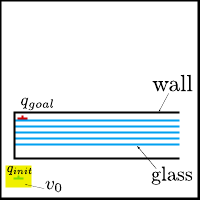} \\
(a) \hspace{0.5cm}  & (b) \hspace{0.5cm} & (c) 
\end{tabular}

\caption{\small Experiment domains: (a) \textsc{OneHallway}, (b) \textsc{HorseshoeHallway}, and (c) \textsc{GlassHallway}.}
\label{fig:domains}
\end{figure}

Throughout our experiments, we focus on domains with a discretized workspace and configuration space. The workspace is planar and the configuration space represents the position of a reference point of the robot, along with the robot's orientation. 
There are 6 possible actions, one in the positive and negative direction for each dimension of the configuration space, forming a 6-connected lattice in the configuration space.
The viewcone is fixed and is about 1.5 times the length of the robot. 

In our FPNNT, we set the maximum lookback size $M=32$ (much smaller than worst-case brute-force lookback $L$, which is on the order of the domain length\footnote{For our experimental domains, the shortest paths grow linearly with respect to the domain length.}, e.g. $10^4$).

\subsection{Experimental Results}


For each of the domains shown in Figure~\ref{fig:domains}, we parameterize the length of the domain for demonstrating asymptotic performance of the use of different visibility query optimizations. For \textsc{OneHallway} and \textsc{HorseshoeHallway}, we vary the vertical lengths from 1000 to 15000 cells, incremented by 1000 cells, while maintaining the widths of the domains. For the \textsc{GlassHallway}, we vary the sizes from $100\times 100$ to $1500\times 1500$ cells, incremented by 100 cells. Note that as we increase the size of the \textsc{GlassHallway}, the number of glass hallways linearly increases. In all of these domains, the hallway widths remain constant.

\begin{figure}
\centering
\begin{subfigure}[b]{0.49\linewidth}
  \includegraphics[width=\linewidth]{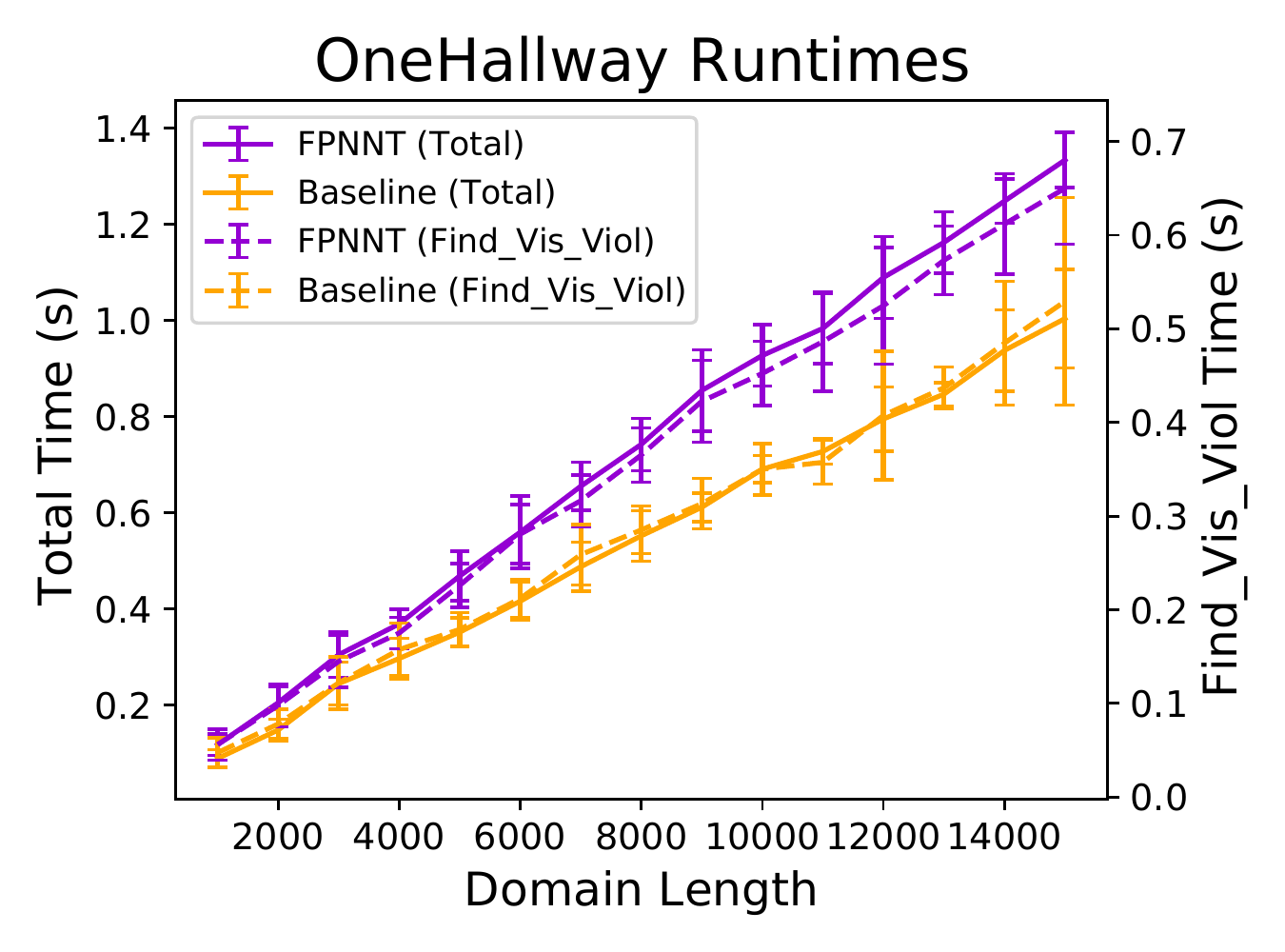}
\end{subfigure}
\begin{subfigure}[b]{0.49\linewidth}
  \includegraphics[width=\linewidth]{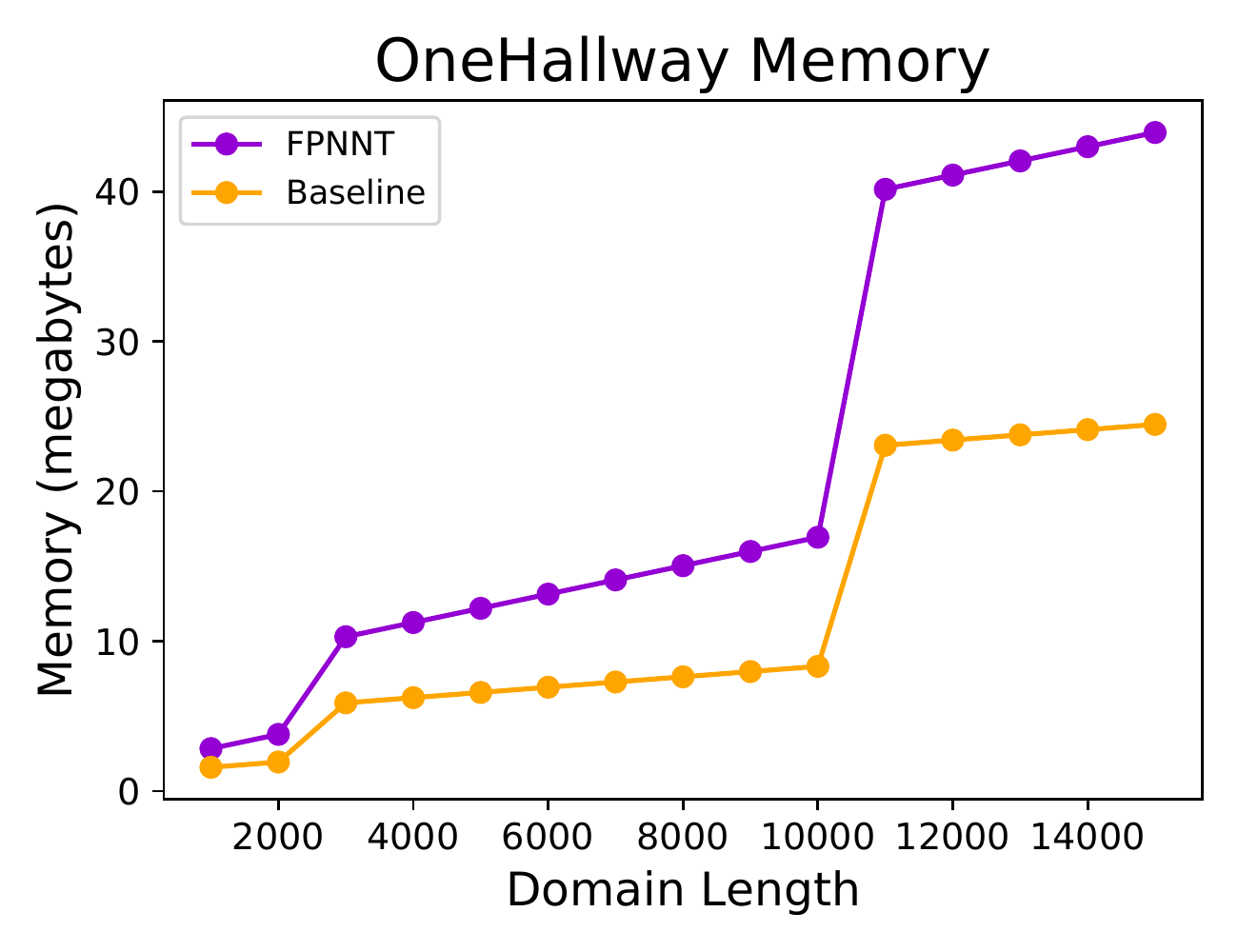}
\end{subfigure}\\

\begin{subfigure}[b]{0.49\linewidth}
  \includegraphics[width=\linewidth]{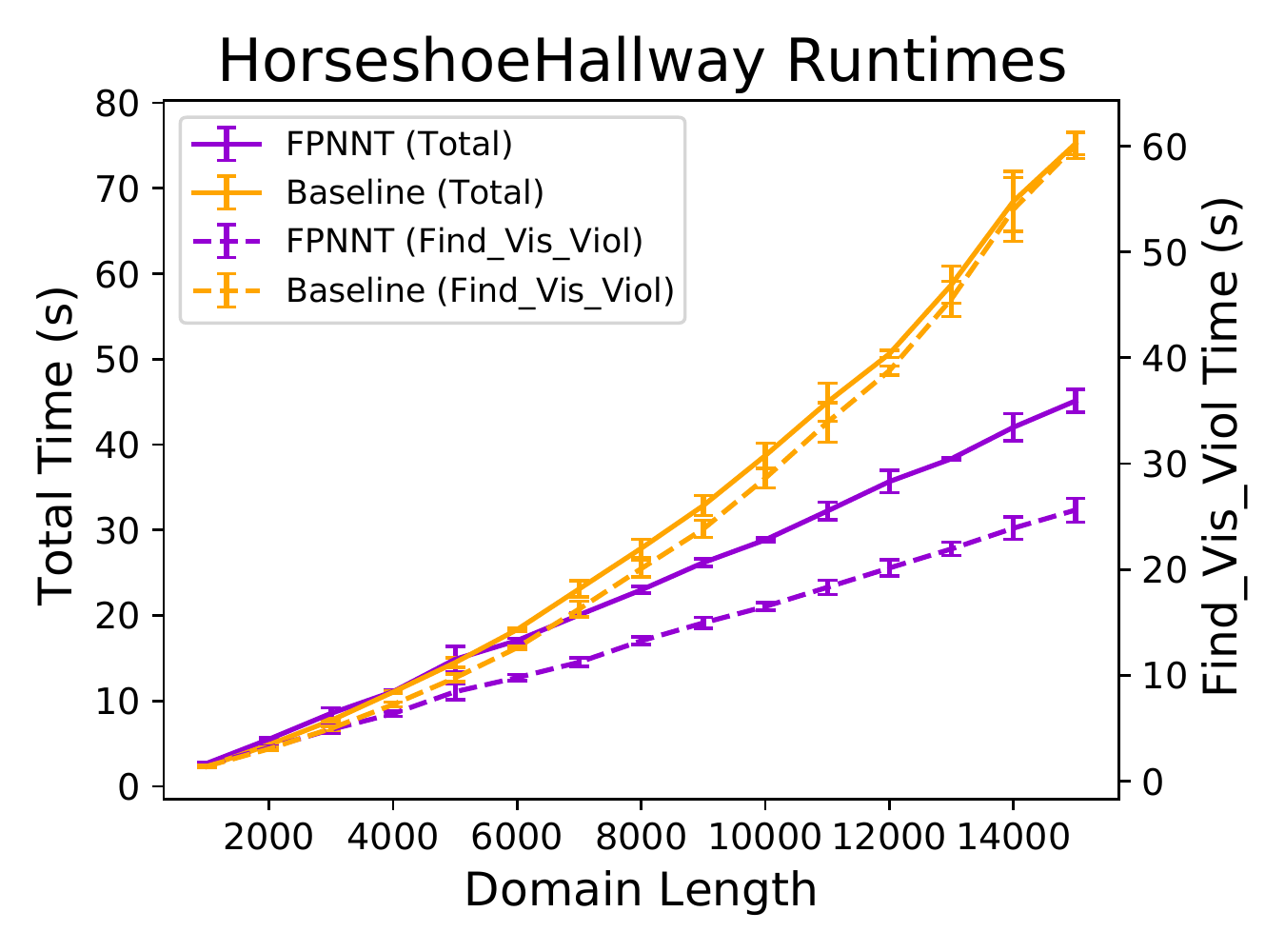}
\end{subfigure}
\begin{subfigure}[b]{0.49\linewidth}
  \includegraphics[width=\linewidth]{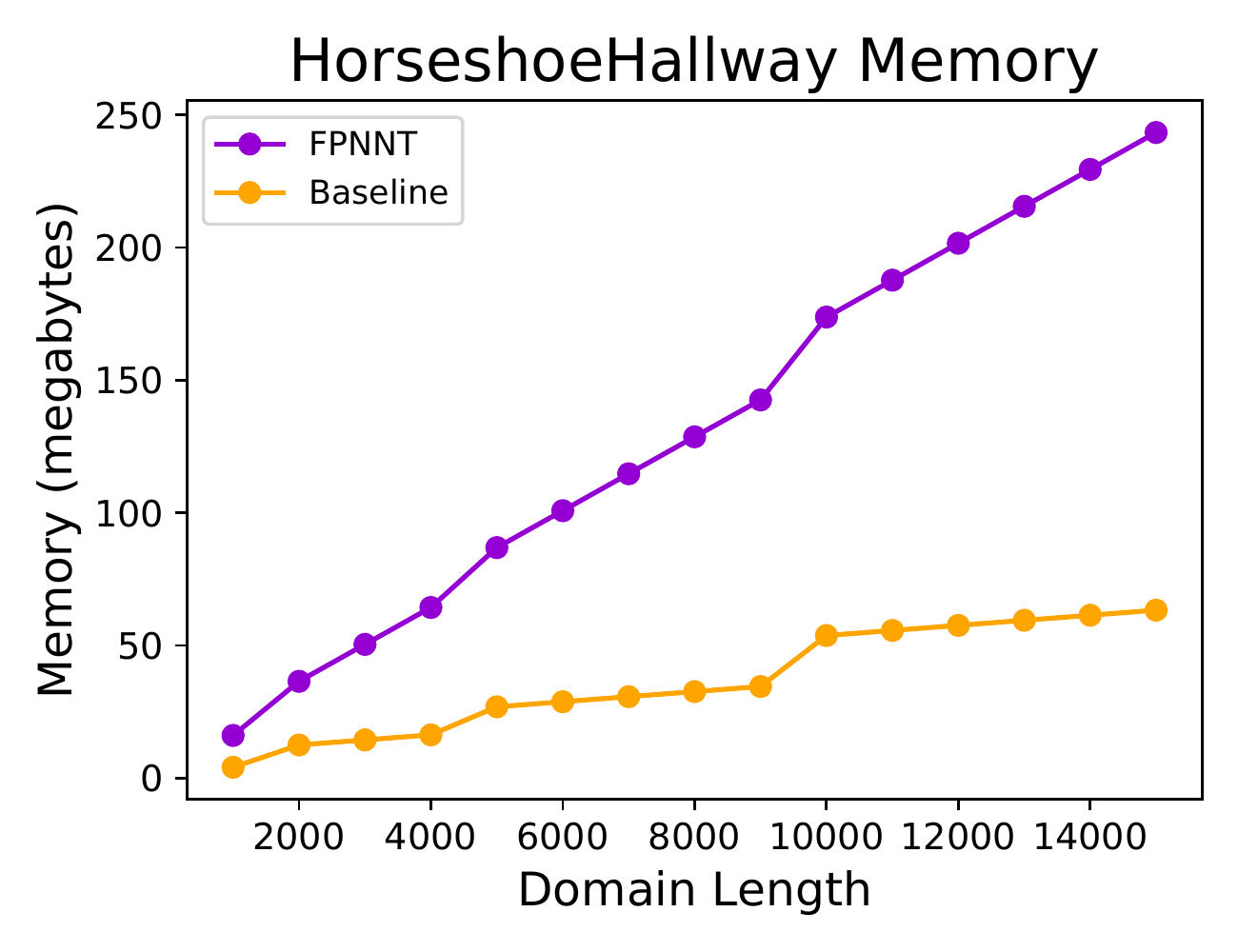}
\end{subfigure}

\caption{\small These plots show the results for the \textsc{OneHallway}  and \textsc{HorseshoeHallway} domains. The left column shows the overall runtimes and total times spent in \textsc{Find\_Vis\_Viol}. The right column shows the memory storage (via \texttt{Base.summarysize}) of the search trees. Discontinuities in memory use as the length grows are attributed to Julia data structure implementation details \cite{JuliaBaseDict}.}
\label{fig:results_runtimes_storage}
\end{figure}

\begin{figure}
\centering
\begin{subfigure}[b]{0.49\linewidth}
  \includegraphics[width=\linewidth]{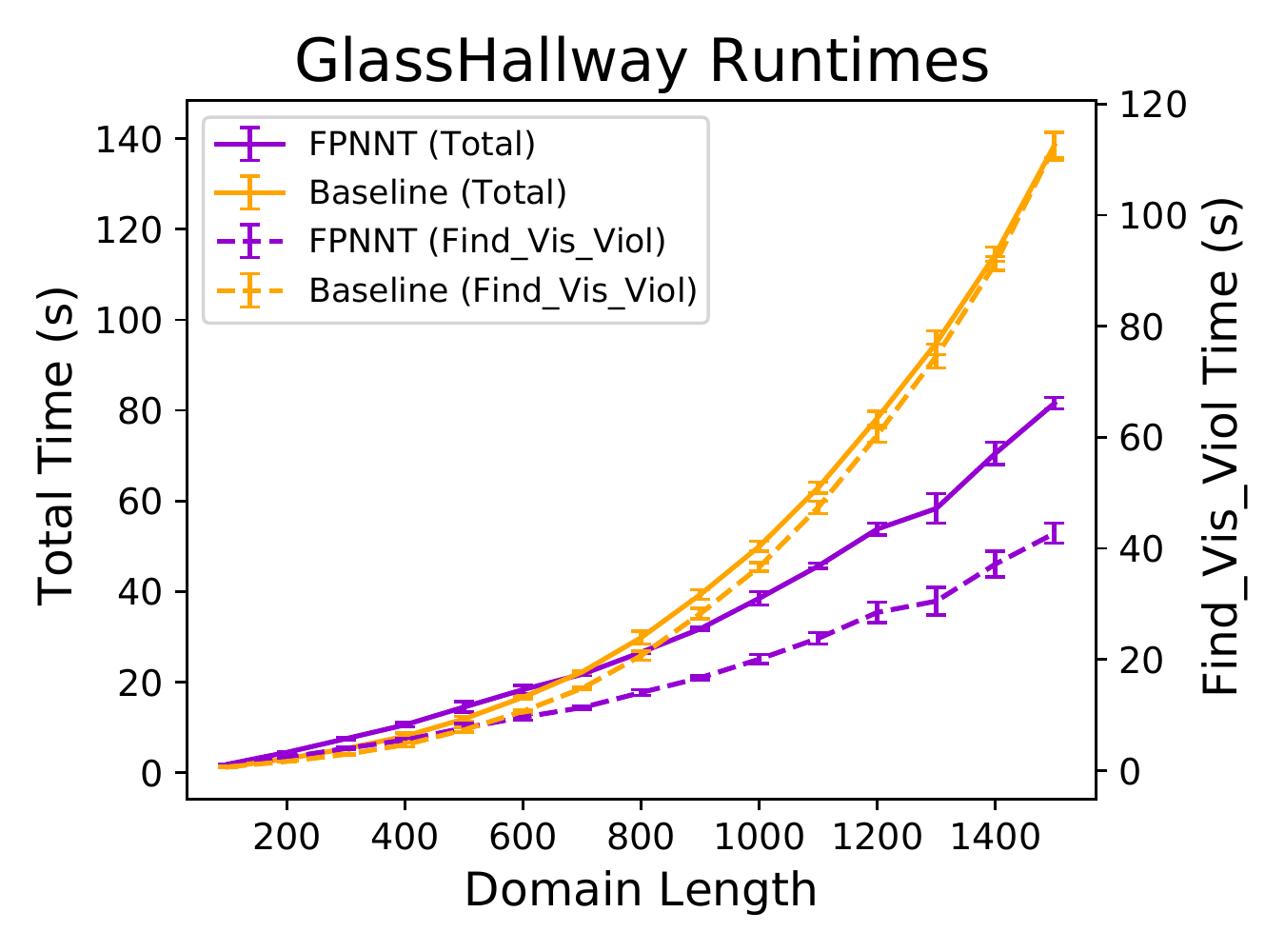}
\end{subfigure}
\begin{subfigure}[b]{0.49\linewidth}
  \includegraphics[width=\linewidth]{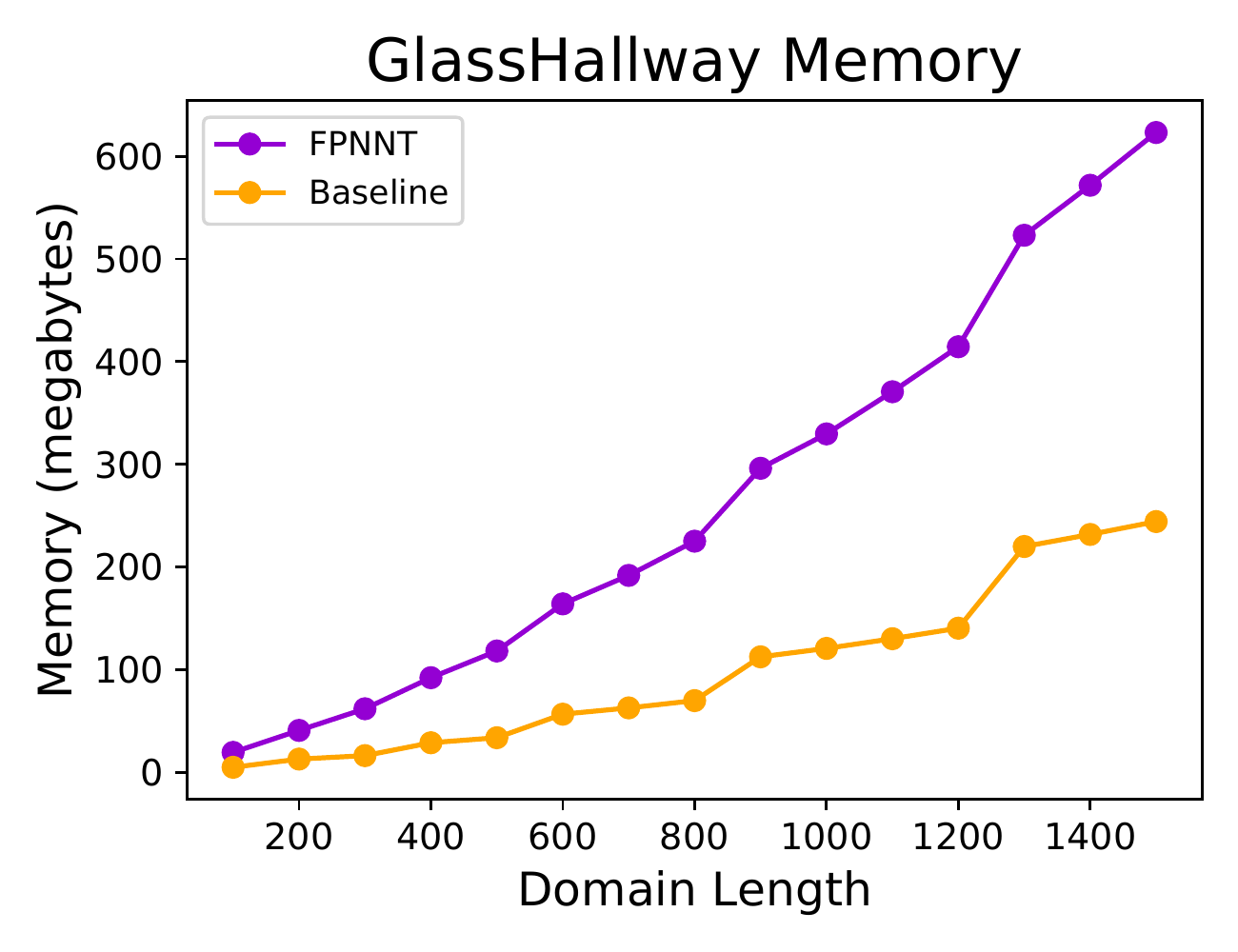}
\end{subfigure}
\caption{\small These plots show similar statistics as shown in Figure~\ref{fig:results_runtimes_storage} but for the \textsc{GlassHallway} domain.}
\label{fig:results_runtimes_storage_glasshallway}
\end{figure}

Figure~\ref{fig:results_runtimes_storage} and Figure~\ref{fig:results_runtimes_storage_glasshallway} show the comparisons of the use of the FPNNT with the baseline in terms of runtimes and search tree storage. We separate the plots for \textsc{GlassHallway} from those corresponding to the other domains to emphasize the difference in scaling domain lengths. The left column of plots shows the comparisons of runtimes and the right column of plots shows the comparisons of search tree storage. For each domain length, we ran 10 iterations of the \textsc{Vamp\_Path\_Vis} algorithm for the baseline and the use of the FPNNT. We collected and reported the average total runtime and the average runtime spent in the \textsc{Find\_Vis\_Viol} function. For each $\vamp$ problem instance and range query method, we reported the overall memory consumed by the search. 

In Figure~\ref{fig:results_runtimes_storage}, we can see that in the simple $\vamp$ problem instance of the \textsc{OneHallway} domain, the total runtime of the search is approximately 1 second for even the highest domain sizes we tested with. We can see that the baseline method outperforms our method in this simple case. 
It is also apparent from the empirical data that \textsc{Find\_Vis\_Viol} takes the majority of the runtime, thus illustrating the importance of these performance improvements.

As the unseen swept region computation requires more visibility calculations, \textsc{HorseshoeHallway} and \textsc{GlassHallway} illustrate a significant performance improvement.
In \textsc{GlassHallway}, we observe the performance improvement at even smaller sizes. The overall runtime improvement does not exactly match the reduction in time spent in \textsc{Find\_Vis\_Viol} due to the FPNNT insertion times, which are already accounted for in the total runtimes. 

The second column of plots in Figure~\ref{fig:results_runtimes_storage} and Figure~\ref{fig:results_runtimes_storage_glasshallway} shows the total memory of the search trees. Since each node of the FPNNT is stored at each node of the search tree, our method is expected to have a search tree that requires more memory than the baseline. Across all of the domains, the search tree storage plots reflect this expectation. However, for the domains tested, the ratio of memory usage between our method and the baseline method remains under 4.2 even as the domain size grows. 

\section{Discussion}
Fully persistent data structures are key to some algorithms \cite{kaplanPersistentDataStructures2017}, and find applications in strictly functional programming, where mutation is prohibited \cite{okasakiPurelyFunctionalData1996}.
To our knowledge, this is the first use of a fully persistent data structure to accelerate planning. We showcase the approach in discrete search on a lattice, but the idea extends to any tree-based approach, such as RRT or searching for paths within a Probabilistic Road Map (PRM).
In the applications we discussed, the FPSDS accelerates what is otherwise a linear-time search to determine points in a range, where these points represent the centers of bounding balls of predetermined radius. For large instances of a recent formulation called Visibility-Aware Motion Planning, we find that the FPNNT improves the planning runtimes by a factor of 2, at the expense of memory use.
Other choices of FPSDS, such as using an R-Tree  \cite{guttmanRtreesDynamicIndex1984} to represent the extent of bounding volumes, may further improve the performance in various settings.
\newline


{
\renewcommand{\baselinestretch}{.5}
\noindent
\textbf{Acknowledgements}:
We gratefully acknowledge support from NSF grant 1723381; from AFOSR grant FA9550-17-1-0165; from ONR grant N00014-18-1-2847; from MIT-IBM Watson Lab and from the MIT Quest for Intelligence.
}

\printbibliography

\end{document}